# DECENTRALIZED SMART CHARGING OF LARGE-SCALE EVS USING ADAPTIVE MULTI-AGENT MULTI-ARMED BANDITS

Sharyal ZAFAR
ENS Rennes – France
sharyal.zafar@ens-rennes.fr

Raphaël FÉRAUD
Orange Labs – France
raphael.feraud@orange.com

Anne BLAVETTE
ENS Rennes - France
anne.blavette@ens-rennes.fr

Guy Camilleri
Paul Sabatier University – France
guy.camilleri@irit.fr

Hamid BEN AHMED
ENS Rennes – France
benahmed@ens-rennes.fr

## ABSTRACT

*The drastic growth of electric vehicles and photovoltaics can introduce new challenges, such as electrical current congestion and voltage limit violations due to peak load demands. These issues can be mitigated by controlling the operation of electric vehicles i.e., smart charging. Centralized smart charging solutions have already been proposed in the literature. But such solutions may lack scalability and suffer from inherent drawbacks of centralization, such as a single point of failure, and data privacy concerns. Decentralization can help tackle these challenges. In this paper, a fully decentralized smart charging system is proposed using the philosophy of adaptive multi-agent systems. The proposed system utilizes multi-armed bandit learning to handle uncertainties in the system. The presented system is decentralized, scalable, real-time, model-free, and takes fairness among different players into account. A detailed case study is also presented for performance evaluation.*

## INTRODUCTION

The increasing adoption of electric vehicles (EVs) and photovoltaic panels (PVs) can significantly reduce the dependence on fossil fuels. This would help reach climate goals by reducing carbon emissions. However, an uncontrolled introduction of these new grid elements may also impact the functioning of the distribution networks. The charging of EVs can be concentrated at certain times, which could lead to issues such as voltage drops, electrical current congestion, and power outages. Utility companies can tackle these challenges through grid reinforcement solutions. However, such solutions may come with higher network investment costs. In recent years, smart charging has been presented as a viable alternative to balance the load on the electrical grid by controlling the charging of electric vehicles.

Smart charging of EVs requires the synergy of different electricity market actors at different levels. The constraints of these different actors (distribution system operator and prosumers) must be satisfied while optimizing the desired objective function of the smart charging optimization problem. A wide variety of smart charging solutions based on centralized or hierarchical architecture have been presented in the literature [1]. These mentioned solutions serve well to optimize the charging of EVs. However, such systems may suffer from issues such as a lack of scalability for real-time operation, potential single point of failure, and privacy concerns over each prosumer's data. Most of the mentioned solutions require an accurate distribution network model, which is not always present. Such challenges can be tackled through the decentralization of the system.

Multi-agent systems (MASs) using the concepts of control theory, graph theory, and game theory had been presented in the literature [1]. In [2], an agent-based control system for the smart charging of EVs is presented. In [3], a heuristic MAS has been presented for the same purpose, but the DSO constraints have not been considered. Standard reinforcement learning (RL) based multi-agent systems have also been proposed to control different smart grid operations [4]. In RL, the goal of each agent is to maximize the running total of its reward, which is obtained through interactions with the environment [4].

In our presented MAS, the concepts of adaptive multi-agent systems (AMAS) have been utilized. In AMAS, cooperation among different system agents is used for self-organization [5]. In [5], Self-organization is defined as, "the mechanism or the process enabling a system to change its organization without explicit external command during execution time." The agents in an AMAS generally hold properties such as autonomy, reactivity, locality, and social ability. This methodology has already been used to tackle problems from different domains [5]. The AMAS theory proposes meta-rules that can be combined with other self-organization approaches (such as reinforcement learning), to further improve performance, especially under stochastic conditions.

As stated earlier, the application of standard reinforcement learning algorithms for smart grid control is not uncommon. However, the choice of the learning algorithm can highly impact the performance of the designed system. Q-learning algorithms with function approximations can suffer from instability, overestimation, and underestimation due to delusional bias in learning [6]. Also, theoretical convergence results for reinforcement learning with functional approximations are still limited. In contrast to standard reinforcement learning, there exists a sub-class, called multi-armed bandit (MAB) [9]. As it is a simpler sub-class of Markov Decision Processes (MDP), it converges faster compared to the most commonly used standard reinforcement learning algorithms (Q-learning or DQN learning). This is a significant advantage of MAB algorithms for smart grid applications (where an Oracle giving the outcome of agent's every action is not available), as the agent is expected to learn online, and thus the total cost of the agent is directly linked to its convergence time. Also, well-defined theoretical guarantees are present for MAB algorithms due to their simpler nature. Systems based on bandit algorithms have been proposed in the literature to control smart grid operations [7], as well as to optimize modern communication networks [8].

The concepts of AMAS and bandit have been combined to propose a fully decentralized adaptive multi-agent system. The AMAS framework results in the decentralization of the system and ensures data privacy. The use of multi-armed bandit learning helps each decision-making agent in the system to handle the uncertainties in the system while performing optimal control of the EVs charging.

## PROBLEM DESCRIPTION

The objective of each EV is to minimize its daily total cost of charging considering the dynamic electricity pricing. It can be written as:

$$obj: min \sum_{i=1}^{m} c(i)P(i)\Delta i \qquad (1)$$

where $c(i)$ is the instantaneous cost of electricity, $P(i)$ is the instantaneous charging power of the EV, $m$ is the total number of decision instants, and $\Delta i$ is the duration of each instant. Furthermore, it is considered that EVs can utilize the generated PV energy without any cost. Hence, learning the daily uncertain PV energy production trend also becomes essential to optimize the daily charging cost.

A set of constraints must be satisfied as well. There should not be any electrical current congestion or voltage limit violations in the distribution network. Furthermore, each EV should be sufficiently charged at its departure time to satisfy the prosumer.

## ADAPTIVE MULTI-AGENT SYSTEM

An AMAS is a special sub-class of MAS where agents cooperate with their neighboring agents. An AMAS agent does not require a full perception of the system as it perceives only a small part of its immediate environment (neighboring agents). In an AMAS, cooperation of agents at microlevel helps solve the problem at macrolevel. Each AMAS agent has its own objective and a set of possible actions. An agent is executed in a loop and goes through three main stages i.e., perception, decision, and action. Each agent perceives the input from the environment through sensors, selects an action to be carried out based on the observed information, and implements the decided action. At each iteration, the agent decides whether to take an action to satisfy its own objective or to help a neighboring agent through cooperation. This decision is made based on the comparison of criticalities. Each agent holds a criticality value (within the range $[-1, 1]$). A criticality value of 0 indicates that the agent is not critical and if it is needed, it can take an action to help a more critical neighboring agent. Let $Cr_p$ be the criticality of the agent $p$, and $[Cr_n]$ be the set of criticalities of its neighboring agents. Then, the objective of $p$ according to the AMAS theory can be written as:

$$obj: min \left( max(Cr_p, [Cr_n]) \right) \qquad (2)$$

i.e., minimize the maximum of the agent's own criticality and its neighboring agents' criticalities.

### Agentification and communication

The agentification process is defined as the modeling of physical components of an electrical distribution network as AMAS agents. An example of this process is shown in Fig. 1. In our proposed AMAS, there are three main types of intelligent agents in the system. Each electrical line, electrical bus, and electric vehicle present in the distribution network has been modeled as a line agent, a bus agent, and an EV agent respectively. The functioning of each agent type is described in the next subsection.

The communication is done through a pre-defined protocol. Each agent requires the perception of only its neighboring agents, which means each AMAS agent communicates only with its neighboring agents. The neighborhood of each agent is defined based on the physical structure of the distribution network. The communication request, sent from one agent to its neighboring agent(s), is in the form of a pair $(Cr_f, [X]_f)$. The first element of this dual pair is the criticality associated with the most critical agent. The second element represents the set of EVs (uniformly sampled from all EVs present in the distribution network), picked by the most critical agent for cooperative actions.

### Agent's functioning

**Line agent**

The objective of each line agent is to avoid electrical current congestion in the system. First, the line agent receives requests from its neighboring agents. Then, each line agent calculates its criticality based on the magnitude of the electrical current. Finally, for cooperation, it finds the most critical request (including itself), and forwards it to its neighboring agents. This functioning is described in Algorithm 1. Terms $[E]$, and $[X]$ represents all the EVs charging from the grid in the distribution network, and the set of EVs picked by the agent (through uniform sampling) to request for cooperative actions. This uniform sampling helps in maintaining fairness among all EV agents.

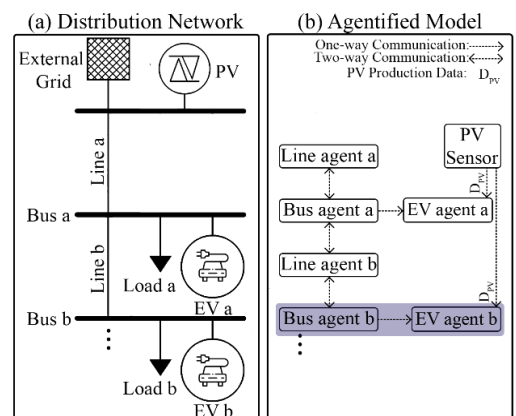

Fig. 1. (a) Example sub-section of a distribution network (b) Agentified AMAS model (the shaded area represents the neighborhood of the EV agent b) of the example network in (a).

| Algorithm 1: Line Agent's Functionality |
|---|
| 1:   **begin agent cycle** |
| 2:     /* *Perception Stage* |
| 3:     $[R] \leftarrow$ neighboring agents' received requests |
| 4:     $I(i) \leftarrow$ Instantaneous electrical line current |
| 5:     $I_{rated} \leftarrow$ Rated electrical line current |
| 6:     /* *Decision stage* |
| 7:     $Cr \leftarrow 1$ **if** $(I(i) > I_{rated})$ **else** $0$ |
| 8:     **if** $(|Cr|>|$neighboring agents' criticalities$|)$ **then** |
| 10:        $Cr_f \leftarrow Cr$ |
| 11:        $[X]_f \leftarrow [X]$ uniformly sampled from $[E]$ |
| 12:    **else** |
| 13:        $Cr_f \leftarrow$ received highest criticality |
| 14:        $[X]_f \leftarrow [X]$ of the most critical request |
| 15:    /* *Action stage* |
| 16:    Send $(Cr_f, [X]_f)$ to all neighboring agents |
| 17:  **end agent cycle** |

**Bus agent**
Each bus agent ensures that the bus voltage remains within the desired limits. It follows the same steps as the line agent. The main difference is that it calculates its criticality based on the bus' over-voltage or under-voltage. Furthermore, priority is given to the line congestion issue over the voltage limit violation issue in the decision step. The functioning of the bus agent is given in Algorithm 2.

| Algorithm 2: Bus Agent's Functionality |
|---|
| 1:   **begin agent cycle** |
| 2:     /* *Perception Stage* |
| 3:     $[R] \leftarrow$ neighboring agents' received requests |
| 4:     $V(i) \leftarrow$ Instantaneous electrical bus voltage |
| 5:     $V_{max} \leftarrow$ Maximum rated electrical bus voltage |
| 6:     $V_{min} \leftarrow$ Minimum rated electrical bus voltage |
| 7:     /* *Decision stage* |
| 8:     $Cr \leftarrow 1$ **if** $(V(i) < V_{min})$ **else** { -1 **if** $(V(i) > V_{max})$ **else** $0$} |
| 9:     **if** (any line congestion request is received) **then** |
| 10:        $Cr_f \leftarrow$ Criticality of the congested line |
| 11:        $[X]_f \leftarrow [X]$ picked by the congested line |
| 12:    **else if** $(Cr \neq 0)$ **then** |
| 13:        $Cr_f \leftarrow Cr$ |
| 14:        $[X]_f \leftarrow [X]$ uniformly sampled from $[E]$ |
| 15:    /* *Action stage* |
| 16:    Send $(Cr_f, [X]_f)$ to all neighboring agents |
| 17:  **end agent cycle** |

**EV agent**
The EV agent is the main decision-making entity in the system, as it decides to charge (or not charge) at a particular instant based on its objective and the received neighborhood criticalities. This decision problem is formulated as a combinatorial multi-armed bandit problem (CMAB). In CMAB, the agent plays a super arm (a combination of base arms with unknown distributions) to observe a reward from the environment. Based on this reward, the estimated return of each played base arm is updated. The goal of each agent is to find the best super arm [9]. In the studied problem, this means, each day $d$ consists of $m\epsilon[m]$ equally spaced instants (base arms), linked to the instantaneous electricity cost $c(i)$. The goal of each EV agent is to find the best instants to charge.

Each EV makes a binary decision (to charge at the rated power or to not charge) for each upcoming instant. Assuming the linear structure of super arms gives the following expected reward for the super arm $S$:

$$\mathbb{E}[r(S)] = S^{*T}\theta \quad (3)$$

where $S^* = \arg\max S^T\theta$, and $\theta \epsilon R^m$ is an unknown parameter. Each EV agent uses linear Thompson Sampling to learn this parameter [10]. The optimal super arm is known when $\theta$ is completely known. Based on the $d$-th day estimation of the unknown vector $\hat{\theta}_d$, the best estimation of the super arm on day $d$ is defined as:

$$S_d = \arg\max S^T \hat{\theta}_d \quad (4)$$

and the pseudo-regret after D days of learning is given as:

$$\mathbb{E}[R(D)] = \sum_{d=1}^{D} S^{*T}\theta - \sum_{d=1}^{D} S_d^T \hat{\theta}_d \quad (5)$$

The reward function depends on the criticalities of the neighboring agents, and the criticality of the EV agent. The criticality of the EV agent is linked to the normalized electricity cost as: $Cr_{EV} = c(i)$. Then, the reward function is defined as:

$$Rew(Cr_{EV}, [Cr_n]) = \begin{cases} -\max([Cr_n]); & if\ ([Cr_n] \neq 0) \\ 1 - Cr_{EV}; & if\ ([Cr_n] = 0) \end{cases} \quad (6)$$

The use of multi-armed bandit learning helps in tackling the uncertainties in the studied decentralized problem. The uncertainty in the choice of super arms of other players is handled through selfish Thompson Sampling [8]. The uncertainty in the freely available PV energy production is handled by learning the vector $(\hat{\emptyset}_d \epsilon R^m)$ representing the instantaneous PV energy production, using Thompson Sampling. For real-time operation, after each instant, each EV agent updates the required number of grid charging instants, based on the latest information, as follows:

$$k_f = \left\lceil \frac{60 E_{bat}(SoC_f - SoC_s)}{\Delta i P_{max}\eta_{chrg}} - \frac{\sum_{i=t_{arrive}}^{t_{depart}} \hat{\emptyset}_i}{P_{max}\eta_{chrg}} \right\rceil - k_p \quad (7)$$

where $k_p$ is the number of already charged instants from the grid. Terms $E_{bat}, P_{max}, SoC_f, SoC_s, \eta_{chrg}, t_{arrive}$, and $t_{depart}$ stands for the energy capacity of the EV battery, the rated charging power of the EV, the desired final SoC, the initial SoC at the connection time, the charging efficiency of the EV, the arrival time, and the departure time of the EV respectively. The function $\lceil . \rceil$ represents the ceiling function. The functioning of the EV agent is given in Algorithm 3. In the algorithm, $\mathbb{1}\{.\}$ stands for the indicator function and $\|.\|_1$ returns the total number of selected base arms. The assumption of linear structured super arms allows evaluation of the best super arm in $O(m)$, which makes the system computationally scalable.

| **Algorithm 3:** EV Agent's Functionality |
|---|
| 1:      $\alpha \epsilon R_+, \beta \epsilon R_+$ |
| 2:      $A, Y \coloneqq I_{m,m}, \hat{\theta}, \hat{\emptyset} \coloneqq 0_m, b, z \coloneqq 0_m$ |
| 3:      **for** d = 1, 2, 3, … **do** |
| 4:         $R, P, M, k_p \coloneqq 0_m$ |
| 5:         $\tilde{\theta} \sim N(\hat{\theta}, \alpha^2 A^{-1}), \tilde{\emptyset} \sim N(\hat{\emptyset}, \beta^2 Y^{-1})$ |
| 6:         **for** i = 1, 2, 3, …, m∀ $t_{arrive} \leq i \leq t_{depart}$ **do** |
| 7:             begin agent cycle |
| 8:             /* *Decision Stage* |
| 9:             Calculate $k_f$ using (7) |
| 10:            Play $S_d$ using (4) $\ni \sum_{l>i,d} S_{l,d} = \|S_d\|_1 = k_f$ |
| 11:            /* *Action stage* |
| 12:            Set EV's instantaneous charging power |
| 13:            /* *Perception stage* |
| 14:            $[R] \leftarrow$ received requests from neighbors |
| 15:            $c(i) \leftarrow$ Instantaneous electricity cost |
| 16:            $R_i \leftarrow$ Instantaneous reward using (6) |
| 17:            $P_i \leftarrow$ Instantaneous PV sensor data |
| 18:            $M_i \coloneqq \mathbb{1}\{i \epsilon S_d\}, k_p \coloneqq \|M_i\|_1$ |
| 19:            end agent cycle |
| 20:         **end for** |
| 21:         $A \coloneqq A + MM^T; Y \coloneqq Y + I_k I_k^T$ |
| 22:         $b \coloneqq b + R; z \coloneqq z + P$ |
| 23:         $\hat{\theta} \coloneqq A^{-1} b; \hat{\emptyset} \coloneqq Y^{-1} z$ |
| 24:      **end for** |

## EXPERIMENTAL EVALUATIONS

### Experimental settings

Two case studies have been performed to evaluate the performance of the proposed AMAS. First, a small-scale case study with 55 households, 55 PVs, and 55 EVs. Second, a large-scale case study with 10,175 households, 10,175 PVs, and 10,175 EVs. The topologies of distribution networks modelled for both studies are shown in Fig. 2. Each sub-district in the shown topologies represents the IEEE low voltage test feeder (LVTF) [11]. The arrival and departure times of the EVs are set based on a real-life dataset [12]. The irradiance data to model the PV generation are obtained through the national renewable energy laboratory (NREL) database [13]. The instantaneous PV generation is defined as:

$$P_{PV}(i) = A \eta_{PV} Irr(i) \quad (8)$$

where $A$ is the area of the PV panels, $\eta_{PV}$ is the efficiency of the PV panels, and $Irr(i)$ is the instantaneous irradiance. Terms $E_{bat}, \eta_{chrg}, P_{max}$, and $SoC_f$ are set to 52

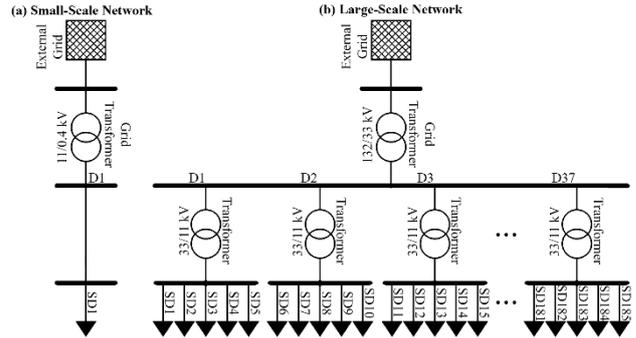

Fig. 2. Topologies of the studied distribution networks (a) Small-scale network (b) Large-scale network.

kWh, 0.95, 7 kW, and 0.8. Furthermore, it is assumed in these studies that the communication speed among agents is much faster than the resolution of each EV's decision instant. This ensures no delays in the communication. To evaluate the fairness level achieved by the optimization strategy, the following equation is used:

$$F([D]) = \frac{1}{1 + \left(\frac{\sigma_{[D]}}{\overline{[D]}}\right)^2} \quad (9)$$

where $[D]$ is the set of per-unit charging costs of each EV, $\sigma_{[D]}$ is the standard deviation of the set $[D]$, and $\overline{[D]}$ is the mean of the set $[D]$. This value ranges between 0 (completely unfair system) and 1 (completely fair system). The following three charging strategies are compared:

**Uncontrolled (basic) charging strategy**
In this charging strategy, the EV starts charging at its maximum power as soon as it is plugged-in for charging. This is a non-optimal charging strategy.

**Centralized optimization charging strategy**
In this strategy, a centralized node performs optimization and decides the charging strategy of each EV present in the distribution network. The detailed explanation of the constraints for the centralized optimization is present in [14]. In this study, it is assumed that a completely accurate PV production forecast is known. This would mean that the solution obtained through the centralized strategy is the optimal solution, and hence can be treated as the lower-bound to better evaluate the performance of our proposed AMAS, which learns the PV production trend.

**Proposed AMAS charging strategy**
This is our proposed charging strategy using the concepts of AMAS combined with multi-armed bandit learning.

### Results

**Small-scale study**
The learning algorithm converges within 30 simulation days. The mean of average learning rewards of all EVs in the network is shown in Fig. 3(a). The results of the next 30 days have been used for performance evaluation. The results are summarized in Fig. 4. It can be seen that the proposed AMAS significantly reduces the system's cost compared to the basic charging strategy. It should be noted that the shown centralized optimal cost is highly unlikely in real-life as it requires an error-free knowledge of the PV energy production forecast. There are also no violations of the constraints in the proposed AMAS, unlike the basic charging strategy. Furthermore, the fairness index value for our proposed AMAS is 0.994, which practically confirms that it takes fairness into account.

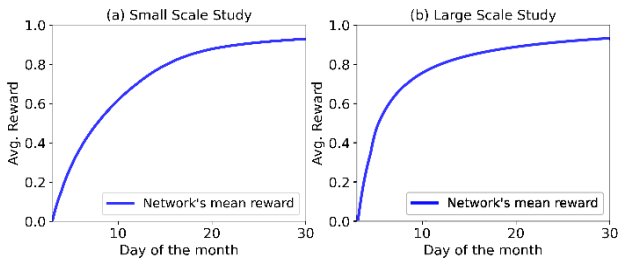

Fig. 3. Average learning reward of the total network (a) Small-scale case study (b) Large-scale case study.

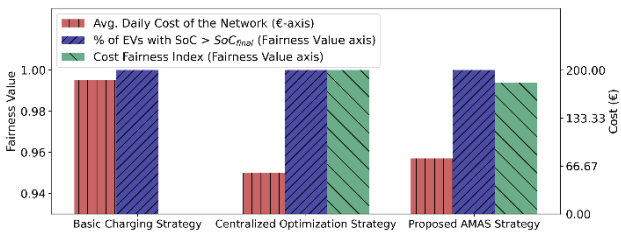

Fig. 4. Results for the small-scale distribution network study

**Large-scale study**
The centralized optimization strategy does not work due to the large number of agents in the system. However, the proposed AMAS still converges, as shown in Fig. 3(b). After convergence, the results of the next 30 days are used to evaluate the performance. The performance summary against the basic charging strategy is presented in Fig. 5. The proposed AMAS results in no constraints violations. However, both current and voltage constraints are violated in the basic charging strategy due to peak load demands. The cost of the system is also reduced significantly if the proposed AMAS smart charging strategy is followed. No fairness index is calculated for the basic charging as it is not an optimization strategy. However, for our proposed AMAS, this value comes out to be 0.993.

## CONCLUSION

An AMAS combined with MAB learning is presented for smart charging. The proposed system follows the AMAS theory for self-organization. The uncertainties are handled through combinatorial multi-armed bandit learning. The detailed performance evaluation confirms that the proposed decentralized system significantly improves the system's performance compared to the basic charging strategy and produces near-optimal results. The impact of using PV forecasts as contextual data in the MAB algorithm can be studied in the future.

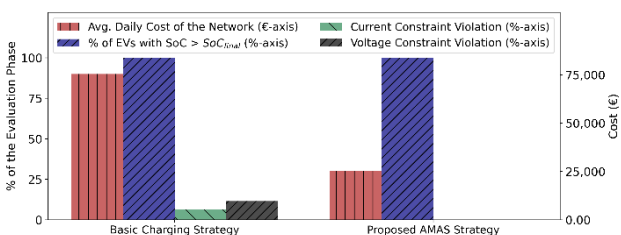

Fig. 5. Results for the large-scale distribution network study